\def\year{2018}\relax
\documentclass[letterpaper]{article} 
\usepackage{aaai18}  
\usepackage{times}  
\usepackage{helvet}  
\usepackage{courier}  
\usepackage{url}  
\usepackage{graphicx}  
\usepackage{subcaption}
\usepackage{color}
\usepackage{amssymb,textcomp}

\usepackage[ruled,noline]{algorithm2e}
\usepackage{amsmath}
\newcommand{\bigo}[1]{\mathcal{O}\left(#1\right)}
\newcommand{\eg}{{\it e.g.}}
\newcommand{\ie}{{\it i.e.}}
\usepackage{pifont}
\newcommand{\norm}[1]{\left\|#1\right\|}
\newcommand{\argmin}{\mathop{\rm argmin}}
\newcommand{\argmax}{\mathop{\rm argmax}}
\newcommand{\st}{\text{subject to}}
\newcommand{\pair}[2]{\langle#1, #2\rangle}
\newcommand{\reals}{{\mbox{$\mathbf{R}$}}}
\newcommand{\quo}[1]{``#1''}

\begin{document}
%
\title{DID: Distributed Incremental Block Coordinate Descent for\\ Nonnegative Matrix Factorization}
\author{Tianxiang Gao, Chris Chu\\
	Department of Electrical and
	Computer Engineering, Iowa State University, Ames, IA, 50011, USA\\
	\{gaotx, cnchu\}@iastate.edu\\
}
\maketitle
\begin{abstract}
Nonnegative matrix factorization (NMF) has attracted much attention in the last decade as a dimension reduction method in many applications. Due to the explosion in the size of data, naturally the samples are collected and stored distributively in local computational nodes. Thus, there is a growing need to develop algorithms in a distributed memory architecture. We propose a novel distributed algorithm, called \textit{distributed incremental block coordinate descent} (DID), to solve the problem. By adapting the block coordinate descent framework, closed-form update rules are obtained in DID. Moreover, DID performs updates incrementally based on the most recently updated residual matrix. As a result, only one communication step per iteration is required. The correctness, efficiency, and scalability of the proposed algorithm are verified in a series of numerical experiments. 
\end{abstract}

\section{Introduction}
Nonnegative matrix factorization (NMF) \cite{Lee:99} extracts the latent factors in a low dimensional subspace. The popularity of NMF is due to its ability to learn \textit{parts}-based representation by the use of nonnegative constraints. Numerous successes have been found in document clustering \cite{xu2004document,lu2017nonconvex}, computer vision \cite{Lee:99}, signal processing \cite{gao2016minimum,lu2017nonconvex}, etc.

Suppose a collection of $N$ samples with $M$ nonnegative measurements is denoted in matrix form $X\in\reals _+^{M\times N}$, where each column is a sample. The purpose of NMF is to approximate $X$ by a product of two nonnegative matrices $B\in\reals_+^{M\times K}$ and $C\in\reals_+^{K\times N}$ with a desired low dimension $K$, where $K\ll \min \{M,N\}$. The columns of matrix $B$ can be considered as a \textit{basis} in the low dimension subspace, while the columns of matrix $C$ are the \textit{coordinates}. NMF can be formulated as an optimization problem in \eqref{opt:nmf}:
\begin{subequations}\label{opt:nmf}
	\begin{eqnarray}
	\underset{B,C}{\text{minimize}}	&&f(B,C)=\frac{1}{2}\norm{X-BC}_F^2\\
	\st && B,C\geq0,
	\end{eqnarray}	
\end{subequations}
where \quo{$\geq0$} means element-wise nonnegative, and $\norm{\cdot}_F$ is the Frobenius norm. The problem \eqref{opt:nmf} is nonconvex with respect to variables $B$ and $C$. Finding the global minimum is NP-hard \cite{vavasis2009complexity}. Thus, a practical algorithm usually converges to a local minimum.

Many algorithms have been proposed to solve NMF such as \textit{multiplicative updates} (MU) \cite{lee2001algorithms}, \textit{hierarchical alternating least square} (HALS) \cite{cichocki2007hierarchical,li2009fastnmf}, \textit{alternating direction multiplier method} (ADMM) \cite{zhang2010alternating}, and \textit{alternating nonnegative least square} (ANLS) \cite{kim2011fast}. Amongst those algorithms, ANLS has the largest reduction of objective value per iteration since it exactly solves \textit{nonnegative least square} (NNLS) subproblems using a \textit{block principal pivoting} (BPP) method \cite{kim2011fast}. Unfortunately, the computation of each iteration is costly. The algorithm HALS, on the other hand, solves subproblems inexactly with cheaper computation and has achieved faster convergence in terms of time \cite{kim2011fast,gillis2012accelerated}. Instead of iteratively solving the subproblems, ADMM obtains closed-form solutions by using auxiliary variables. A drawback of ADMM is that it is sensitive to the choice of the tuning parameters, even to the point where poor parameter selection can lead to algorithm divergence \cite{sun2014alternating}.

Most of the proposed algorithms are intended for centralized implementation, assuming that the whole data matrix can be loaded into the RAM of a single computer node. In the era of massive data sets, however, this assumption is often not satisfied, since the number of samples is too large to be stored in a single node. As a result, there is a growing need to develop algorithms in distributed system. Thus, in this paper, we assume the number of samples is so large that the data matrix is collected and stored distributively. Such applications can be found in e-commerce (\eg, Amazon), digital content streaming (\eg, Netflix) \cite{koren2009matrix} and technology (\eg, Facebook, Google) \cite{tan2016faster}, where they have hundreds of millions of users. 

Many distributed algorithms have been published recently. The distributed MU \cite{liu2010distributed,yin2014scalable} has been proposed as the first distributed algorithm to solve NMF. However, MU suffers from slow and ill convergence in some cases \cite{lin2007convergence}. \citeauthor{kannan2016high} \cite{kannan2016high} proposed high performance ANLS (HPC-ANLS) using 2D-grid partition of a data matrix such that each node only stores a submatrix of the data matrix. Nevertheless, six communication steps per iteration are required to obtain intermediate variables so as to solve the subproblems. Thus, the communication overhead is significant. Moreover, the computation is costly as they use ANLS framework. The most recent work is distributed HALS (D-HALS) \cite{zdunek2017distributed}. However, they assume the factors $B$ and $C$ can be stored in the shared memory of the computer nodes, which may not be the case if $N$ is large. \citeauthor{boyd2011distributed} \cite{boyd2011distributed} suggested that ADMM has the potential to solve NMF distributively. \citeauthor{du2014maxios} \cite{du2014maxios} demonstrated this idea in an algorithm called Maxios. Similar to HPC-ANLS, the communication overhead is expensive, since every latent factor or auxiliary variable has to be gathered and broadcasted over all computational nodes. As a result, eight communication steps per iteration are necessary. In addition, Maxios only works for sparse matrices since they assume the whole data matrix is stored in every computer node.
	
In this paper, we propose a distributed algorithm based on block coordinate descent framework. The main contributions of this paper are listed below.

\begin{itemize}
	\item We propose a novel distributed algorithm, called \textit{distributed incremental block coordinate descent} (DID). By splitting the columns of the data matrix, DID is capable of updating the coordinate matrix $C$ in parallel. Leveraging the most recent residual matrix, the basis matrix $B$ is updated distributively and incrementally. Thus, only one communication step is needed in each iteration. 
	\item A scalable and easy implementation of DID is derived using \textit{Message Passing Interface} (MPI). Our implementation does not require a master processor to synchronize. 
	\item Experimental results showcase the correctness, efficiency, and scalability of our novel method. 
\end{itemize}

The paper is organized as follows. In Section 2, the previous works are briefly reviewed. Section 3 introduces a distributed ADMM for comparison purpose. The novel algorithm DID is detailed in Section 4. In Section 5, the algorithms are evaluated and compared. Finally, the conclusions are drawn in Section 6.

\section{Previous Works}

In this section we briefly introduce three standard algorithms to solve NMF problem \eqref{opt:nmf}, \ie, ANLS, HALS, and ADMM, and discuss the parallelism of their distributed versions.

\paragraph{Notations.}Given a nonnegative matrix $X\in\reals_+^{M\times N}$ with $M$ rows and $N$ columns, we use $x_i^r\in\reals_+^{1\times N}$ to denote its $i$-th row, $x_j\in\reals_+^{M\times 1}$ to denote the $j$-th column, and $x_{ij}\in\reals_+$ to denote the entry in the $i$-th row and $j$-th column. In addition, we use $x_i^{rT}\in\reals_+^{N\times 1}$ and $x_j^T\in\reals_+^{1\times M}$ to denote the transpose of $i$-th row and $j$-th column, respectively.

\subsection{ANLS}
The optimization problem \eqref{opt:nmf} is biconvex, \ie, if either factor is fixed, updating another is in fact reduced to a \textit{nonnegative least square} (NNLS) problem. Thus, ANLS \cite{kim2011fast} minimizes the NNLS subproblems with respect to $B$ and $C$, alternately. The procedure is given by
\begin{subequations}\label{opt:ANLS}
	\begin{eqnarray}
	&C:=\argmin_{C\geq 0}\norm{X-BC}_F^2\\
	&B:=\argmin_{B\geq 0}\norm{X-BC}_F^2.
	\end{eqnarray}	
\end{subequations}
The optimal solution of a NNLS subproblem can be achieved using BPP method. 

A naive distributed ANLS is to parallel $C$-minimization step in a column-by-column manner and $B$-minimization step in a row-by-row manner. HPC-ANLS \cite{kannan2016high} divides the matrix $X$ into 2D-grid blocks, the matrix $B$ into $P_r$ row blocks, and the matrix $C$ into $P_c$ column blocks so that the memory requirement of each node is $\mathcal{O}(\frac{MN}{P_rP_c}+ \frac{MK}{P_r}+\frac{NK}{P_c})$, where $P_r$ is the number of rows processor and $P_c$ is the number of columns processor such that $P=P_cP_r$ is the total number of processors. To really perform updates, the intermediate variables $CC^T$, $XC^T$, $B^TB$, and $B^TX$ are computed and broadcasted using totally six communication steps. Each of them has a cost of $\log P\cdot(\alpha + \beta\cdot NK)$, where $\alpha$ is \textit{latency}, and $\beta$ is \textit{inverse bandwidth} in a \textit{distributed memory network} model \cite{chan2007collective}. The analysis is summarized in Table \ref{tab:an}.


\subsection{HALS}
Since the optimal solution to the subproblem is not required when updating one factor, a comparable method, called HALS, which achieves an approximate solution is proposed by \cite{cichocki2007hierarchical}. The algorithm HALS successively updates each column of $B$ and row of $C$ with an optimal solution in a closed form.

The objective function in \eqref{opt:nmf} can be expressed with respect to the $k$-th column of $B$ and $k$-th row of $C$ as follows
\begin{equation*}
\resizebox{0.95\hsize}{!}{$
\Big\|X-BC\Big\|_F^2=\Big\| X-\sum_{i=1}^K b_ic_i^r \Big\|=\Big\|X-\sum_{i\neq k}b_ic_i^r-b_kc_k^r\Big\|_F^2
$},
\end{equation*}
Let $A\triangleq X-\sum_{i\neq k}b_ic_i^r$ and fix all the variables except $b_{k}$ or $c_k^r$. We have subproblems in $b_k$ and $c_k^r$
\begin{subequations}
	\begin{align}
		&\underset{b_{k}\geq 0}{\min} \; \norm{A-b_kc_k^r}_F^2, \\
		&\underset{c_k^r\geq 0}{\min} \; \norm{A-b_kc_k^r}_F^2
	\end{align}
\end{subequations}
By setting the derivative with respect to $b_k$ or $c_k^r$ to zero and projecting the result to the nonnegative region, the optimal solution of $b_k$ and $c_k^r$ can be easily written in a closed form
\begin{subequations}
\begin{align}
b_{k}&:=\left[(c_k^rc_k^{rT})^{-1}(Ac_k^{rT})\right]_+\\
c_{k}^r&:=\left[(b_k^Tb_k)^{-1}(A^Tb_k)\right]_+
\end{align}
\end{subequations}
where $\left[z\right]_+$ is $\max\{0, z\}$. Therefore, we have $K$ inner-loop iterations to update every pair of $b_k$ and $c_k^r$. With cheaper computational cost, HALS was confirmed to have faster convergence in terms of time.

\citeauthor{zdunek2017distributed} in \citeyear{zdunek2017distributed} proposed a distributed version of HALS, called DHALS. They also divide the data matrix $X$ into 2D-grid blocks. Comparing with HPC-ANLS, the resulting algorithm DHALS only requires two communication steps. However, they assume matrices $B$ and $C$ can be loaded into the shared memory of a single node. Therefore, DHALS is not applicable in our scenario where we assume $N$ is so big that even the latent factors are stored distributively. See the detailed analysis in Table 1.

\subsection{ADMM}
The algorithm ADMM \cite{zhang2010alternating} solves the NMF problem by alternately optimizing the Lagrangian function with respect to different variables. Specifically, the NMF \eqref{opt:nmf} is reformulated as
\begin{subequations}\label{opt:admm}
	\begin{eqnarray}
	\underset{B,C,W,H}{\text{minimize}}	&&\frac{1}{2}\norm{X-WH}_F^2\\
	\st && B=W, C=H\\
		&& B,C\geq0,
	\end{eqnarray}	
\end{subequations}
where $W\in\reals^{M\times K}$ and $H\in\reals^{K\times N}$ are \textit{auxiliary variables} without \textit{nonnegative} constraints. The \textit{augmented Lagrangian function} is given by
\begin{multline}
\mathcal{L}(B,C,W,H;\Phi,\Psi)_\rho=\frac{1}{2}\norm{X-WH}_F^2+\pair{\Phi}{B-W}\\+\pair{\Psi}{C-H}+\frac{\rho}{2}\norm{B-W}_F^2 + \frac{\rho}{2}\norm{C-H}_F^2
\end{multline}
where $\Phi\in\reals^{M\times K}$ and $\Psi\in\reals^{K\times N}$ are \textit{Lagrangian multipliers}, $\pair{\cdot}{\cdot}$ is the matrix inner product, and $\rho>0$ is the penalty parameter for equality constraints. By minimizing $\mathcal{L}$ with respect to $W$, $H$, $B$, $C$, $\Phi$, and $\Psi$ one at a time while fixing the rest, we obtain the update rules as follows
\begin{subequations}
	\begin{align}
		W &:= (XH^T+\Phi+\rho B)(HH^T+\rho I_K)^{-1}\\
		H &:= (W^TW+\rho I_K)^{-1}(W^TX+\Psi+\rho C)\\
		B &:= \left[W-\Phi/\rho\right]_+\\
		C &:= \left[H-\Psi/\rho\right]_+\\
		\Phi&:= \Phi+\rho(B-W)\\
		\Psi&:= \Psi+\rho(C-H)
	\end{align}
\end{subequations}
where $I_K\in\reals^{K\times K}$ is the identity matrix. The auxiliary variables $W$ and $H$ facilitate the minimization steps for $B$ and $C$. When $\rho$ is small, however, the update rules for $W$ and $H$ result in unstable convergence \cite{sun2014alternating}. When $\rho$ is large, ADMM suffers from a slow convergence. Hence, the selection of $\rho$ is significant in practice. 

Analogous to HPC-ANLS, the update of $W$ and $B$ can be parallelized in a column-by-column manner, while the update of $H$ and $C$ in a row-by-row manner. Thus, Maxios \cite{du2014maxios} divides matrix $W$ and $B$ in column blocks, and matrix $H$ and $C$ in row blocks. However, the communication overhead is expensive since one factor update depends on the others. Thus, once a factor is updated, it has to be broadcasted to all other computational nodes. As a consequence, Maxios requires theoretically eight communication steps per iteration and only works for sparse matrices. Table 1 summarizes the analysis.

\begin{table*}[th]
	\resizebox{\textwidth}{!}{
			\begin{tabular}{ |c|r|r|r|r| }
				\hline
				Algorithm & Runtime& Memory per processor & Communication time& Communication volume\\ \hline
				HPC-ANLS & BPP & $\bigo{MN/(P_cP_r)+MK/P_r+NK/P_c}$ & $3(\alpha+\beta NK)\log P_r+3(\alpha+\beta MK)\log P_c$&$\bigo{MKP_c+NKP_r}$\\ \hline
				D-HALS & $\bigo{MNK(1/P_c+1/P_r)}$ & $\bigo{MN/(P_cP_r)+MK+NK}$ & $(\alpha+\beta NK)\log P_r+(\alpha+\beta MK)\log P_c$&$\bigo{MKP_c+NKP_r}$\\ \hline
				Maxios & $\bigo{K^3+MNK/P}$ & $\bigo{MN}$ & $4(2\alpha + \beta (N+M)K)\log P$& $\bigo{(M+N)KP}$\\ \hline
				DADMM & BPP & $\bigo{MN/P+MK}$ & $(\alpha + \beta MK)\log P$& $\bigo{MKP}$\\ \hline
				DBCD & $\bigo{MNK/P}$ & $\bigo{MN/P+MK}$ & $K(\alpha + \beta  MK)\log P$& $\bigo{MKP}$\\ \hline
				DID & $\bigo{MNK/P}$ & $\bigo{MN/P+MK}$ & $(\alpha + \beta MK)\log P$ & $\bigo{MKP}$\\ \hline
			\end{tabular}
	 }
\caption{Analysis of distributed algorithms per iteration on runtime, memory storage, and communication time and volume.}\label{tab:an}
\end{table*}
\section{Distributed ADMM}\label{sec:DADMM}
This section derives a \textit{distributed ADMM} (DADMM) for comparison purpose. DADMM is inspired by another centralized version in \cite{boyd2011distributed,hajinezhad2016nonnegative}, where the update rules can be easily carried out in parallel, and is stable when $\rho$ is small.

As the objective function in \eqref{opt:nmf} is \textit{separable} in columns, we divide matrices $X$ and $C$ into column blocks of $P$ parts
\begin{equation}
\frac{1}{2}\norm{X-BC}_F^2=\sum_{i=1}^P \frac{1}{2}\norm{X_i-BC_i}_2^2,
\end{equation}
where $X_i\in \reals_+^{M\times N_i}$ and $C_i\in \reals_+^{K\times N_i}$ are column blocks of $X$ and $C$ such that $\sum_{i=1}^{P}N_i=N$. 
Using a set of auxiliary variables $Y_i\in\reals^{M\times N_i}$, the NMF \eqref{opt:nmf} can be reformulated as
\begin{subequations}\label{opt:admm2}
	\begin{eqnarray}
	\underset{Y_i,B,C}{\text{minimize}}	&&\sum_{i=1}^P\frac{1}{2}\norm{X_i-Y_i}_F^2\\
	\st && Y_i=BC_i,\quad\text{for } i=1,2,\cdots, P\\
	&& B,C\geq0.
	\end{eqnarray}	
\end{subequations}
The associated augmented Lagrangian function is given by
\begin{multline}
\mathcal{L}(Y_i,B,C;\Lambda_i)_\rho=\sum_{i=1}^P\frac{1}{2}\norm{X_i-Y_i}_F^2\\+\sum_{i=1}^P\pair{\Lambda_i}{Y_i-BC_i}+\sum_{i=1}^P\frac{\rho}{2}\norm{Y_i-BC_i}_F^2,
\end{multline}
where $\Lambda_i\in\reals^{M\times K}$ are the Lagrangian multipliers. The resulting ADMM is
\begin{subequations}
	\begin{align}
	Y_i&:=\argmin_{Y_i}\frac{1}{2}\norm{X_i-Y_i}_2^2+\frac{\rho}{2}\norm{\Lambda_i/\rho+Y_i-BC_i}_F^2\\
	C_i&:=\argmin_{C_i\geq0}\norm{\Lambda_i/\rho+Y_i-BC_i}_2^2\\
	B&:=\argmin_{B\geq 0}\norm{\Lambda/\rho + Y-BC}_F^2\label{up:B}\\
	\Lambda_i&:=\argmax_{\Lambda_i}\pair{\Lambda_i}{Y_i-BC_i}
	\end{align}
\end{subequations} 
where $\Lambda\triangleq [\Lambda_1\;\Lambda_2\cdots \Lambda_P]$ and $Y\triangleq [Y_1\;Y_2\cdots Y_P]$. Clearly, the $Y_i$ update has a closed-form solution by taking the derivate and setting it to zero, \ie,
\begin{align}
	Y_i:=\frac{1}{1+\rho}(X_i -\Lambda_i + \rho BC_i)
\end{align}
Moreover, the updates for $Y_i$, $C_i$, and $\Lambda_i$ can be carried out in \textit{parallel}. Meanwhile, $B$ needs a central processor to update since the step \eqref{up:B} requires the whole matrices $Y$, $C$, and $\Lambda$. If we use the solver BPP, however, we do not really need to gather those matrices, because the solver BPP in fact does not explicitly need $Y$, $C$, and $\Lambda$. Instead, it requires two intermediate variables $W\triangleq CC^T$ and $H\triangleq (\Lambda/\rho+Y)C^T$, which can be computed as follows:
\begin{subequations}
	\begin{align}
		W&\triangleq CC^T=\sum_{i=1}^P C_iC_i^T,\\
		H&\triangleq (\Lambda/\rho+Y)C^T = \sum_{i=1}^P (\Lambda_i/\rho+Y_i)C_i^T.
	\end{align}
\end{subequations} 
It is no doubt that those intermediate variables can be calculated distributively. Let $U_i = \Lambda_i/\rho$, which is called \textit{scaled dual variable}. Using the scaled dual variable, we can express DADMM in a more efficient and compact way. A simple MPI implementation of algorithm DADMM on each computational node is summarized in Algorithm \ref{alg:DADMM}.

\begin{algorithm}
	\KwIn{$X_i$, $C_i$, $B$}
	\textbf{Initialize} $P$ processors, along with $Y_i$, $B$, $C_i$, $X_i$\\
	\Repeat{stopping criteria satisfied}{
			\nl $U_i:=U_i+(Y_i-BC_i)$\\
			\nl $Y_i:=\frac{1}{1+\rho}(X_i-\rho U_i+\rho BC_i)$\\
			\nl $C_i:=\argmin_{C_i\geq0}\norm{U_i+Y_i-BC_i}_2^2$\\
			\nl $(W,H):=Allreduce(C_iC_i^T, (U_i+Y_i)C_i^T)$\label{DADMM:comm}\\
			\nl $B:=\text{BPP}(W,H)$
	}
	\caption{DADMM for each computational node}\label{alg:DADMM}
\end{algorithm}
At line \ref{DADMM:comm} in Algorithm \ref{alg:DADMM}, theoretically we need a master processor to \textit{gather} $C_iC_i^T$ and $(U_i+Y_i)C_i^T$ from every local processor and then \textit{broadcast} the updated value of $CC^T$ and $(U+Y)C^T$ back. As a result, the master processor needs a storage of $\bigo{MKP}$. However, we use a collaborative operation called \textit{Allreduce} \cite{chan2007collective}. Leveraging it, the master processor is discarded and the storage of each processor is reduced to $\bigo{MK}$. 


\section{Distributed Incremental Block Coordinate Descent}
The popularity of ADMM is due to its ability of carrying out subproblems in parallel such as DADMM in Algorithm \ref{alg:DADMM}. However, the computation of ADMM is costly since it generally involves introducing new auxiliary variables and updating dual variables. The computational cost is even more expensive as it is required to find optimal solutions of subproblems to ensure convergence. In this section, we will propose another distributed algorithm that adapts block coordinate descent framework and achieves approximate solutions at each iteration. Moreover, leveraging the current residual matrix facilitates the update for matrix $B$ so that columns of $B$ can be updated incrementally.
 
\subsection{Distributed Block Coordinate Descent}
We firstly introduce a naive parallel and distributed algorithm, which is inspired by HALS, called \textit{distributed block coordinate descent} (DBCD). Since the objective function in \eqref{opt:nmf} is separable, the matrix $X$ is partitioned by columns, then each processor is able to update columns of $C$ in parallel, and prepare messages concurrently to update matrix $B$.

Analogous to DADMM, the objective function in \eqref{opt:nmf} can be expanded as follows
\begin{align*}
\resizebox{\hsize}{!}{$
\Big\|X-BC\Big\|_F^2 =\sum_{j=1}^{N} \Big\|x_j-Bc_j\Big\|^2=\sum_{j=1}^{N}\Big\|x_j-\sum_{k=1}^{K}b_kc_{kj}\Big\|^2
$}
\end{align*}
By coordinate descent framework, we only consider one element at a time. To update $c_{ij}$, we fix the rest of variables as constant, then the objective function becomes
\begin{align}
\sum\nolimits_{j=1}^{N}\norm{x_j-\sum\nolimits_{k\neq i}b_kc_{kj}- b_ic_{ij}}^2.\label{obj:dbcd}
\end{align}
Taking the partial derivative of the objective function \eqref{obj:dbcd} with respect to $c_{ij}$ and setting it to zero, we have
\begin{align}
b_i^T\left(b_ic_{ij}- \left(x_j-\sum\nolimits_{k\neq i}b_kc_{kj}\right)\right) = 0. 
\end{align}
The optimal solution of $c_{ij}$ can be easily derived in a closed form as follows
\begin{subequations}
	\begin{align} 
	c_{ij} :=& \left[\frac{b_i^T(x_j-\sum_{k\neq i}b_kc_{kj})}{b_i^Tb_i}\right]_+\\
	=&\left[\frac{b_i^T(x_j-Bc_j + b_ic_{ij})}{b_i^Tb_i}\right]_+\\
	=&\left[c_{ij} + \frac{b_i^T(x_j-Bc_j)}{b_i^Tb_i}\right]_+\label{sol:c_ij}
	\end{align}
\end{subequations}
Based on the equation \eqref{sol:c_ij}, the $j$-th column of $C$ is required so as to update $c_{ij}$. Thus, updating a column $c_j$ has to be sequential. However, the update can be executed in parallel for all $j$'s. Therefore, the columns of matrix $C$ can be updated independently, while each component in a column $c_j$ is optimized in sequence.

The complexity of updating each $c_{ij}$ is $\bigo{MK}$. Thus, the entire complexity of updating matrix $C$ is $\bigo{MNK^2/P}$. This complexity can be reduced by bringing $x_j-Bc_j$ outside the loop and redefining as $e_j\triangleq  x_j-Bc_j$. The improved update rule is
\begin{subequations}
	\begin{align}
	e_j &:= e_j + b_ic_{ij}\\
	c_{ij} &:= \left[\frac{b_i^Te_j}{b_i^Tb_i}\right]_+\\
	e_j &:= e_j - b_ic_{ij}
	\end{align}\label{up:c}
\end{subequations}
By doing so, the complexity is reduced to $\bigo{MNK/P}$.

The analogous derivation can be carried out to update the $i$-th column of matrix $B$, \ie, $b_i$. By taking partial derivative of the objective function \eqref{obj:dbcd} with respect to $b_i$ and setting it to zero, we have equation
\begin{align}
\sum_{j=1}^{N}\left(b_ic_{ij}-\left(x_j-\sum\nolimits_{k\neq i}b_kc_{kj}\right)\right)c_{ij}=0
\end{align}
Solving this linear equation gives us a closed-form to the optimal solution of $b_i$
\begin{subequations}
	\begin{align}
	b_i:=&\left[\frac{\sum_{j=1}^{N}(x_j-Bc_j+b_ic_{ij})c_{ij}}{\sum_{j=1}^{N}c_{ij}^2}\right]_+\label{sol:b_i 2}\\
	=&\left[b_i+\frac{\sum_{j=1}^{N}(x_j-Bc_j)c_{ij}}{\sum_{j=1}^{N}c_{ij}^2}\right]_+\label{sol:b_i 3}\\
	=&\left[b_i + \frac{(X-BC)c_i^{rT}}{c_i^rc_i^{rT}}\right]_+\label{sol:b_i 4}
	\end{align}
\end{subequations}
Unfortunately, there is no way to update $b_i$ in parallel since the equation \eqref{sol:b_i 4} involves the whole matrices $X$ and $C$. That is the reason why sequential algorithms can be easily implemented in the shared memory but cannot directly be applied in distributed memory. Thus, other works \cite{kannan2016high,zdunek2017distributed,du2014maxios} either use \textit{gather} operations to collect messages from local processors or assume small size of the latent factors.

By analyzing the equation \eqref{sol:b_i 2}, we discover the potential parallelism. We define a vector $y_j$ and a scaler $z_j$ as follows 
\begin{subequations}
	\begin{align}
		y_{j}&\triangleq (x_j-Bc_j+b_ic_{ij})c_{ij}=(e_j+b_ic_{ij})c_{ij}\\
		z_j  &\triangleq  c_{ij}^2
	\end{align}
\end{subequations}
The vector $y_j$ and scaler $z_j$ can be computed in parallel. After receiving messages including $y_j$'s and $z_j$'s from other processors, a master processor updates the column $b_i$ as a scaled summation of $y_{j}$ with scaler $z \triangleq \sum_{j=1}^Nz_j$, that is,
\begin{align}
b_i := \left[y/z\right]_+ 
\end{align}
where $y \triangleq  \sum\nolimits_{j=1}^{N}y_{j}$. Thus, the update for matrix $B$ can be executed in parallel but indirectly. The complexity of updating $b_i$ is $\bigo{MN/P}$ as we reserve error vector $e_j$ and concurrently compute $y_j$ and $z_j$. The complexity of updating entire matrix $B$ is $\bigo{MNK/P}$.

By partitioning the data matrix $X$ by columns, the update for matrix $C$ can be carried out in parallel. In addition, we identify vectors $y_j$'s and scalars $z_j$'s to update matrix $B$, and their computation can be executed concurrently among computational nodes. A MPI implementation of this algorithm for each processor is summarized in Algorithm \ref{alg:DBCD}. 
\begin{algorithm}
	\KwIn{$x_j$, $c_j$, $B$}
	\Repeat{stopping criteria satisfied}{
		\tcp{Update $C$}
		$e_j:= x_j-Bc_j$\\
		\For {all $i\in\{1,2,\cdots,K\}$}{
			Update $c_{ij}$ using equations \eqref{up:c}
		}
		\tcp{Update $B$}
		\For{all $i\in\{1,2,\cdots,K\}$} {
			$e_j=e_j + b_ic_{ij}$\\
			$y_{j}=e_jc_{ij}$\\
			$z_j = c_{ij}^2$\\
			$(y,z)=Allreduce(y_{j}, z_{j})$\\
			$b_i := \left[y/z\right]_+$\\
			$e_j=e_j - b_ic_{ij}$
		}
	}
	\caption{DBCD for each computational node}\label{alg:DBCD}
\end{algorithm}

\subsection{Incremental Update for $b_i$}
The complexity of algorithm DBCD is $\bigo{MNK/P}$ per iteration, which is perfectly parallelizing a sequential block coordinate descent algorithm. However, the performance of DBCD could be deficient due to the delay in network. In principle, DBCD sends totally $KP$ messages to a master processor per iteration, which is even more if we implement DBCD using \textit{Allreduce}. Any delay of a message could cause a diminished performance. In contrast, the algorithm DID has a novel way  to update matrix $B$ incrementally using only a single message from each processor per iteration. 

To successfully update matrix $B$, the bottleneck is to iteratively compute $y_{j}$ and $z_j$ for associated $b_i$ since once $b_i$ is updated, the $y_{j}$ and $z_j$ have to be recomputed due to the change occurred in matrix $B$ from equation \eqref{sol:b_i 3}. Nevertheless, we discovered this change can be represented as several arithmetic operations. Thus, we in fact do not need to communicate every time in order to update each $b_i$.

Suppose that after $t$-th iteration, the $i$-th column of matrix $B$ is given, \ie, $b_i^t$, and want to update it to $b_i^{t+1}$. Let $E=X-BC$, which is the most current residual matrix after $t$-th iteration. From equation \eqref{sol:b_i 4}, we have 
\begin{align}
b_i^{t+1} :=\left[b^t_i+\frac{Ec_i^{rT}}{c_i^rc_i^{rT}}\right]_+\label{eq:b_i^{t+1}}
\end{align}
Once we update $b_i^t$ to $b_i^{t+1}$, we need to update $b_i$ in matrix $B$ so as to get new $E$ to update the next column of $B$, \ie, $b_{i+1}$. However, we do not really need to recalculate $E$. Instead, we can update the value by 
\begin{align}
	E := E + b_i^tc_i^r -  b_i^{t+1}c_i^r
\end{align}
We define and compute a variable $\delta b_i$ as 
\begin{align}
\delta b_i\triangleq b_i^{t+1}-b_i^t.
\end{align}
Using the vector $\delta b_i$, we have a compact form to update $E$
\begin{align}
	E := E -  \delta b_ic_i^r
\end{align}
The updated $E$ is substituted into the update rule of $b_{i+1}$ in equation \eqref{eq:b_i^{t+1}}, and using $b_{i+1}^t$ we obtain
\begin{subequations}
\begin{align}
	b_{i+1}^{t+1} :=&\left[b^t_{i+1}+\frac{(E -  \delta b_ic_i^r)c_{i+1}^{rT}}{c_{i+1}^rc_{i+1}^{rT}}\right]_+\\
	=&\left[b^t_{i+1}+\frac{Ec_{i+1}^{rT}}{c_{i+1}^rc_{i+1}^{rT}} - \frac{c_i^rc_{i+1}^{rT}}{c_{i+1}^rc_{i+1}^{rT}} \delta b_i\right]_+\label{eq:b_i+1}
\end{align}
\end{subequations}
In the equation \eqref{eq:b_i+1}, the first two terms is the same as general update rule for matrix $B$ in DBCD, where $Ec_{i+1}$ can be computed distributively in each computational node. On the other hand, the last term allows us to update the column $b_{i+1}$ still in a closed form but without any communication step. Therefore, the update for matrix $B$ can be carried out incrementally and the general update rule is given by
\begin{align}
	b_{i}^{t+1} :=
	&\left[b^t_{i}+\frac{Ec_i^{rT}}{c_{i}^rc_{i}^{rT}} - \frac{\sum_{k<i}(c_i^rc_k^{rT})\delta b_k}{c_{i}^rc_{i}^{rT}} \right]_+
\end{align}
Comparing to the messages used in DBCD, \ie, $(y_j, z_j)$, we need to compute the coefficients for the extra term, that is, $c_i^rc_k^{rT}$ for all $k<i$. Thus, a message communicated among processors contains two parts: the weighted current residual matrix $W_j$, and a lower triangular matrix $V_j$ maintaining the inner product of matrix $C$. The matrices $W_j$ and $V_j$ are defined as below
\begin{align}
	W_j&\triangleq \begin{bmatrix} 
	\vert & \vert & \cdots & \vert\\
	e_jc_{1j} & e_jc_{2j} & \cdots & e_jc_{Kj}\\
	\vert & \vert & \cdots & \vert
	\end{bmatrix}\label{W_j}
	\\
	V_j&\triangleq \begin{bmatrix} 
	c_{1j}^2 		& 0 			& 0 			& \cdots & 0\\
	c_{2j}c_{1j} 	& c_{2j}^2 		& 0 			& \cdots & 0\\
	\vdots 			& \vdots 		& \ddots 		& \vdots & 0\\
	c_{Kj}c_{1j} 	& c_{Kj}c_{2j} 	& c_{Kj}c_{3j} 	& \cdots &c_{Kj}^2
	\end{bmatrix}\label{V_j}
\end{align}
Using variables $W_j$ and $V_j$, the update rule to columns of matrix $B$ becomes
\begin{align}
	b_i := \left[b_i + w_i/v_{ii}-\sum_{k<i}(v_{ik}/v_{ii})\delta b_k\right]_+
\end{align}
where $w_i$ is the $i$-th column of matrix $W$, $v_{ij}$ is the $i$-th component of $j$-th column of matrix $V$, and matrices $W$ and $V$ are the summations of matrices $W_j$ and $V_j$, respectively, \ie, $W \triangleq  \sum\nolimits_{j=1}^{N}W_j$ and $V \triangleq  \sum\nolimits_{j=1}^{N}V_j$.


For each processor, they store a column of $X$, a column of $C$, and the matrix $B$. They execute the same algorithm and a MPI implementation of this incremental algorithm for each computational node is summarized in Algorithm \ref{alg:DID}. Clearly, the entire computation is unchanged and the volume of message stays the same as DBCD, but the number of communication is reduced to once per iteration.

\begin{algorithm}
	\KwIn{$x_j$, $c_j$, $B$}
	\Repeat{stopping criteria satisfied}{
		\tcp{Update $C$}
		$e_j:= x_j-Bc_j$\\
		\For {all $i\in\{1,2,\cdots,K\}$}{
			Update $c_{ij}$ using equations \eqref{up:c}
		}
		Compute $W_j$ and $V_j$ from equations \eqref{W_j} and \eqref{V_j}.\\
		$(W,V):=Allreduce(W_j, V_j)$\\
		\tcp{Update $B$}
		\For{all $i\in\{1,2,\cdots,K\}$} {
			$b_i^{t+1} := \left[b_i^t + w_i/v_{ii}-\sum_{k<i}(v_{ik}/v_{ii})\delta b_k\right]_+$\\
			$\delta b_i := b_{i}^{t+1}-b_i^t$
		}
	}
	\caption{DID for each computational node}\label{alg:DID}
\end{algorithm}

\section{Experiments}
\begin{figure*}[ht]
	\centering
	\begin{minipage}{.3\linewidth}
		\includegraphics[width=\textwidth]{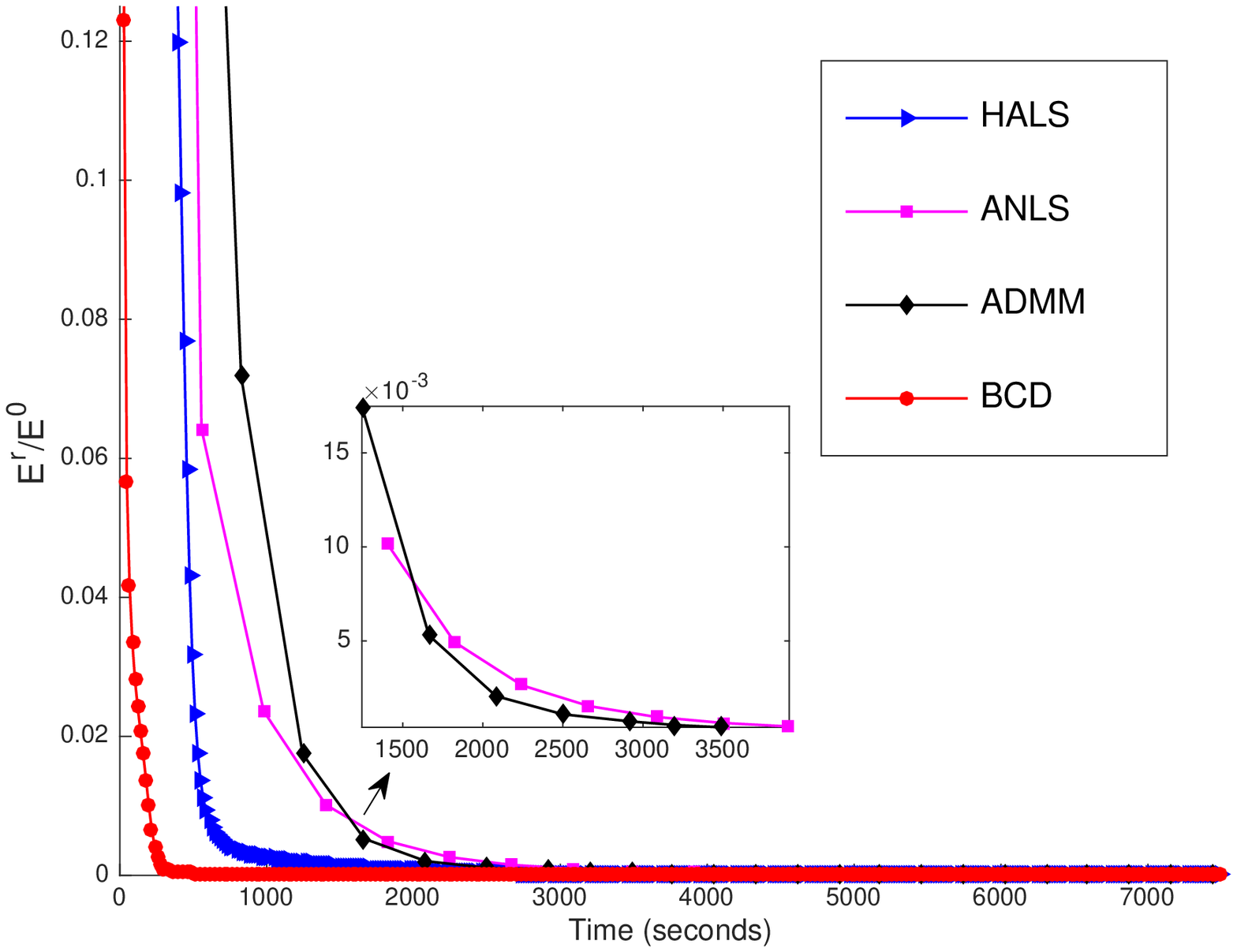}
		\caption*{(a) Sequential}
	\end{minipage}%
	\begin{minipage}{.3\linewidth}
		\includegraphics[width=\textwidth]{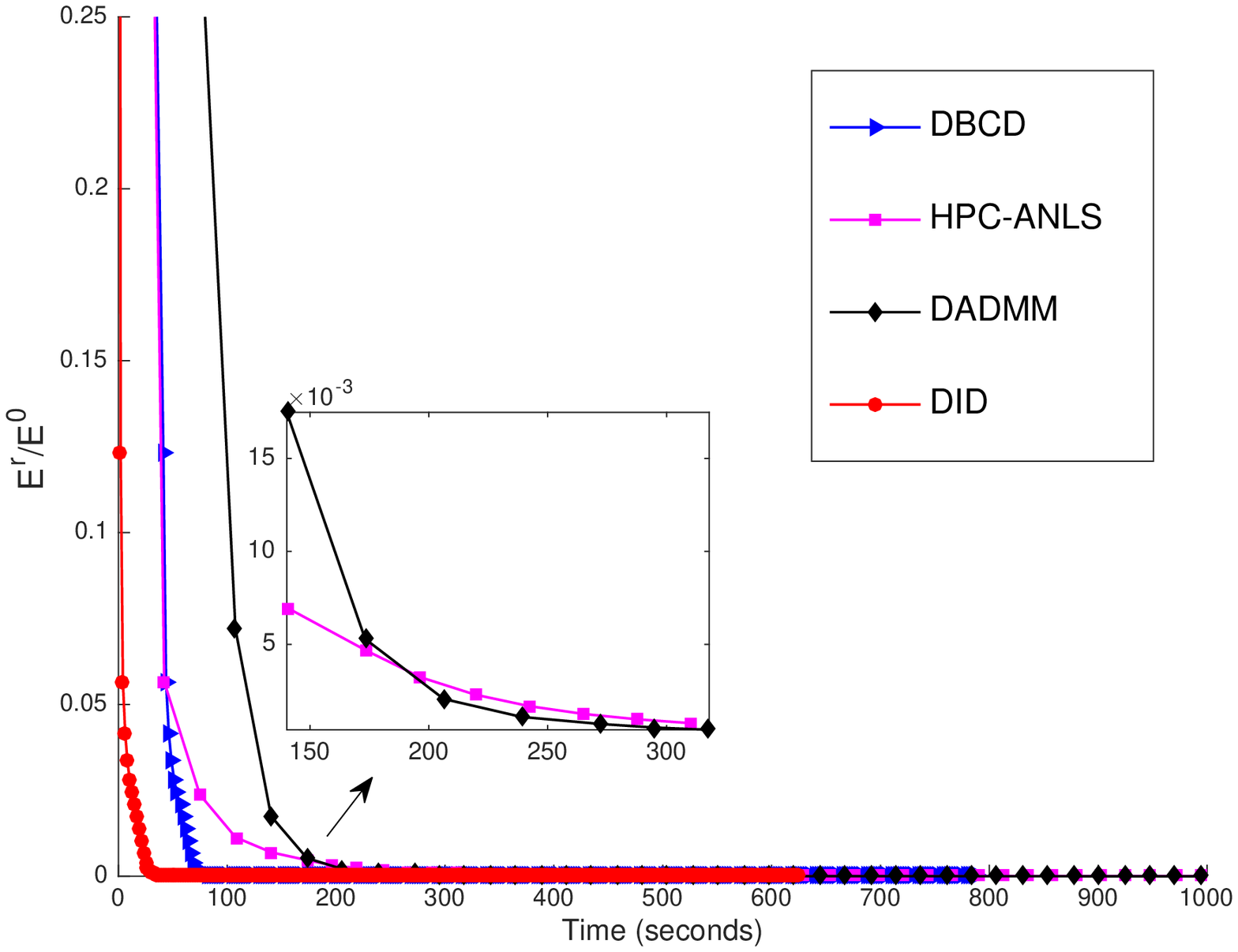}
		\caption*{(b) Distributed}
	\end{minipage}
	\begin{minipage}{.3\linewidth}
		\includegraphics[width=\textwidth]{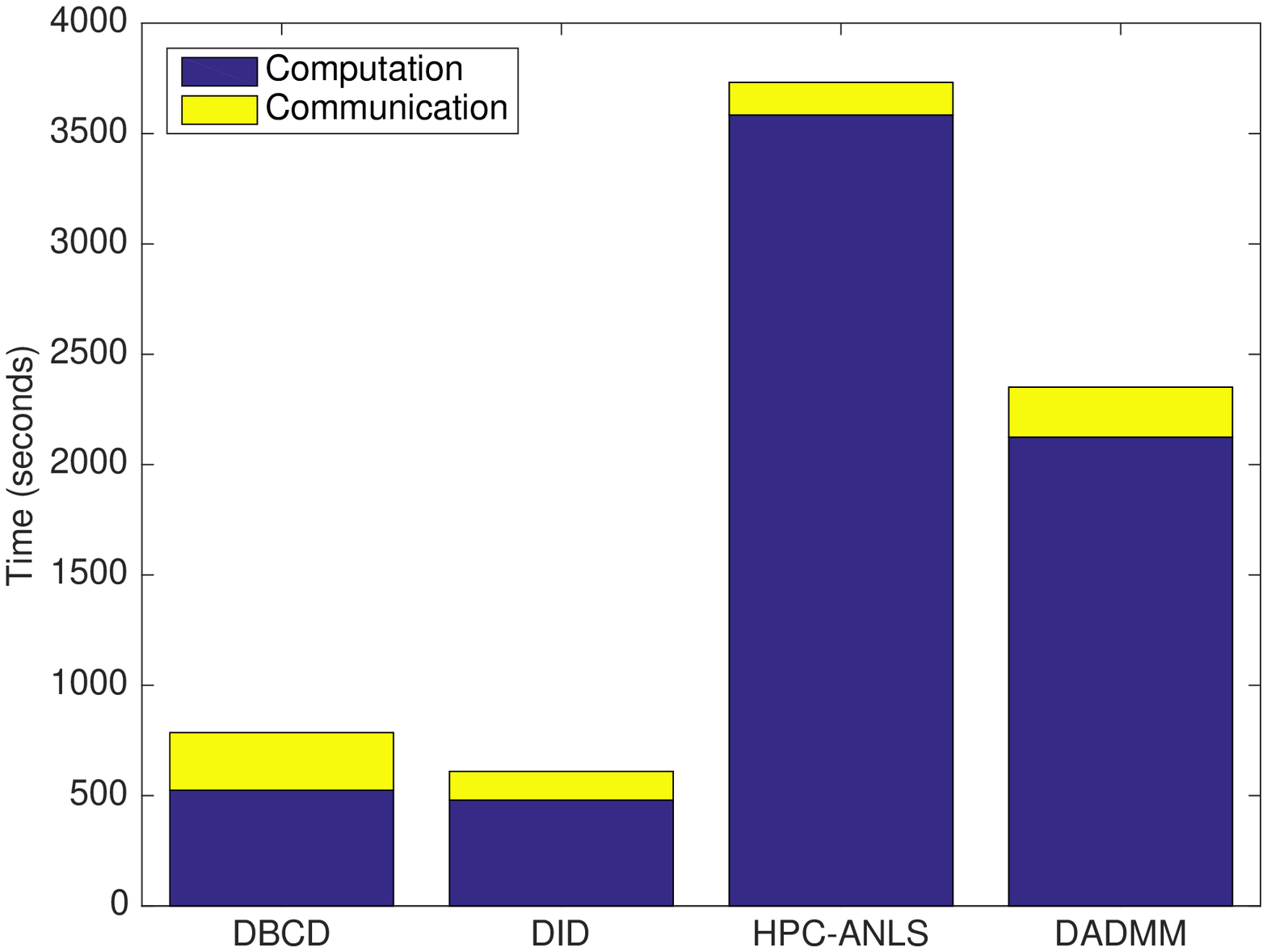}
		\caption*{(c) Computation v.s. communication}
	\end{minipage}
	\caption{Convergence behaviors of different algorithms with respect to time consumption of communication and computation on the dataset with $N=10^8$ samples.}
	\label{fig:conv}
\end{figure*} 
\begin{table*}[ht]
	\resizebox{\textwidth}{!}{
			\centering
			\begin{tabular}{|c|r|r|r|r|r|r|r|r||r|r|r|r|r|r|r|r|}
				\hline
				&	\multicolumn{8}{|c||}{Number of iterations}	&	\multicolumn{8}{|c|}{Time (seconds)}\\ \hline
				$N$& HALS& ANLS & ADMM & BCD  &HPC-ANLS &  DADMM & DBCD&\textbf{DID}& HALS& ANLS & ADMM & BCD &  HPC-ANLS& DADMM&DBCD &\textbf{DID}  \\ \hline
				$10^5$&1281& \textbf{141} & 170 & 549  & \textbf{141} & 170& 549&549 & 16.88& 56.59 & 45.88 & 10.61 & 4.31 &3.46& 1.42& \textbf{1.17}\\ \hline
				$10^6$&225& 238 & \textbf{115} & 396 & 238 & \textbf{115}&396& 396 & 36.86& 630.43 & 476.83 & 95.24  &50.47 & 37.06& 14.04& \textbf{8.61}\\ \hline
				$10^7$&\textbf{596}& 1120 & 1191 & 654 & 1120 & 1191&654 &654 & 587.47& 29234.61 & 31798.51 & 909.76 & 2372.47& 2563.47& 126.01& \textbf{106.60} \\ \hline
				$10^8$&339& 163 & \textbf{97} & 302  &163 & \textbf{97}&302& 302 & 3779.11& 43197.12 & 27590.16 & 8808.92  & 10172.55& 5742.37& 785.57&\textbf{610.09}\\ \hline \hline
				MNIST&495 &\textbf{197} &199 &492 &\textbf{197} &\textbf{199} &492 &492& 705.32& 395.61& 610.65&942.68& \textbf{31.84} & 46.17&170.65 &133.50\\ \hline
				20News&302& \textbf{169}& \textbf{169}& 231& \textbf{169}& \textbf{169}& 231& 231& 2550.02& 745.28& 714.61& 2681.49& \textbf{131.12}& 172.69& 651.52&559.70\\ \hline
				UMist&677& 1001& 953& \textbf{622}& 1001& 953 & \textbf{622}& \textbf{622}& 314.72& 657.14& 836.76& 422.11& 492.72& 471.01& 92.49& \textbf{82.34}\\ \hline
				YaleB&1001& 352& \textbf{224}& 765& 352& \textbf{224}& 765& 765& 223.58& 201.22& 149.35& 236.13& 50.69& 40.61& 44.08&\textbf{36.45}\\ \hline
			\end{tabular}
 }
\caption{Performance comparison for algorithms on synthetic and real datasets with $P=16$ number of computing nodes.}\label{tab:result}
\end{table*}

We conduct a series of numerical experiments to compare the proposed algorithm DID with HALS, ALS, ADMM, BCD, DBCD, DADMM, and HPC-ANLS. The algorithm BCD is the sequential version of DBCD. Due to the ill convergence of ADMM and Maxios in \cite{zhang2010alternating,du2014maxios}, we derive DADMM in Section \ref{sec:DADMM} and set $\rho=1$ as default. Since we assume $M$ and $K$ are much smaller than $N$, HPC-ANLS only has column partition of the matrix $X$, \ie, $P_c=P$ and $P_r=1$.

We use a cluster\footnote{http://www.hpc.iastate.edu/} that consists of 48 SuperMicro servers each with 16 cores, 64 GB of memory, GigE and QDR (40Gbit) InfiniBand interconnects. The algorithms are implemented in C code. The linear algebra operations use GNU Scientific Library (GSL) v2.4\footnote{http://www.gnu.org/software/gsl/} \cite{gough2009gnu}. The Message Passing Interface (MPI) implementation OpenMPI v2.1.0\footnote{https://www.open-mpi.org/} \cite{gabriel04:_open_mpi} is used for communication. 
Note that we do not use multi-cores in each server. Instead, we use \textit{a single core per node} as we want to achieve consistent communication overhead between cores.

Synthetic datasets are generated with number of samples $N=10^5,10^6,10^7$ and $10^8$. Due to the storage limits of the computer system we use, we set the dimension $M=5$ and low rank $K=3$, and utilize $P=16$ number of computational nodes in the cluster. The random numbers in the synthetic datasets are generated by the Matlab command \texttt{rand} that are uniformly distributed in the interval $[0,1]$. 

We also perform experimental comparisons on four real-world datasets. The MNIST dataset\footnote{http://yann.lecun.com/exdb/mnist/} of handwritten digits has 70,000 samples of 28x28 image. The 20News dataset\footnote{http://qwone.com/\texttildelow jason/20Newsgroups/} is a collection of 18,821 documents across 20 different newsgroups with totally 8,165 keywords. The UMist dataset\footnote{https://cs.nyu.edu/\texttildelow roweis/data.html} contains 575 images of 20 people with the size of 112x92. The YaleB dataset\footnote{http://www.cad.zju.edu.cn/home/dengcai/Data/FaceData.html}includes 2,414 images of 38 individuals with the size of 32x32. The MNIST and 20News datasets are sparse, while UMist and YaleB are dense.

The algorithms HALS, (D)BCD, and DID could fail if $\norm{b_i}$ or $\norm{c_i^r}$ is close to zero. This could appear if $B$ or $C$ is badly scaled. That means the entries of $E=X-BC$ are strictly negative. We avoid this issue by using well scaled initial points for the synthetic datasets and $K$-means method to generate the initial values for the real datasets. All the algorithms are provided with the same initial values.

When an iterative algorithm is executed in practice, a stopping criteria is required. In our experiments, the stopping criteria is met if the following condition is satisfied
\begin{align}
	\norm{E^t}^2_F \leq \epsilon \norm{E^0}^2_F,
\end{align}
where $E^t$ is the residual matrix after $t$-th iteration. Throughout the experiments, we set $\epsilon=10^{-6}$ as default. In addition, we combine the stopping criterion with \textit{a limit on time} of $24$ hours and \textit{a maximum iteration} of 1000 for real datasets. The experimental results are summarized in the Table \ref{tab:result}. 

\paragraph{Correctness}In principle, the algorithms HALS, (D)BCD, and DID have the same update rules for the latent factors $B$ and $C$. The difference is the update order. The algorithm DID has the exact same number of iterations as BCD and DBCD, which demonstrates the correctness of DID.

\paragraph{Efficiency}As presented in Table \ref{tab:result}, DID always converges faster than the other algorithms in term of time. HALS and BCD usually use a similar number of iterations to reach the stopping criteria. ANLS and ADMM use much fewer iterations to converge. Thanks to auxiliary variables, ADMM usually converges faster than ANLS. Figure \ref{fig:conv}(a) shows that comparing with HALS, BCD actually reduces the objective value a lot at the beginning but takes longer to finally converge. Such phenomenon can also be observed in the comparison between ANLS and ADMM. In Figure \ref{fig:conv}(b), DID is faster than DBCD. The reason is shown in Figure \ref{fig:conv}(c) that DID involves much less communication overhead than DBCD. Based on the result in Table \ref{tab:result}, DID is about 10-15\% faster than DBCD by incrementally updating matrix $B$. (HPC-)ANLS works better in MNIST and 20News datasets because these datasets are very sparse.



\paragraph{Scalability} As presented in Table \ref{tab:result}, the runtime of DID scales linearly as the number of samples increases, which is much better than the others. It can usually speed up a factor of at least $10$ to BCD using $16$ nodes. (D)ADMM is also linearly scalable, which is slightly better than (HPC-)ANLS. Due to the costly computation, (D)ADMM is not preferred to solve NMF problems.

\section{Conclusion}
In this paper, we proposed a novel distributed algorithm DID to solve NMF in a distributed memory architecture. Assume the number of samples $N$ to be huge, DID divides the matrices $X$ and $C$ into column blocks so that updating the matrix $C$ is perfectly distributed. Using the variables $\delta b$, the matrix $B$ can be updated distributively and incrementally. As a result, only a single communication step per iteration is required. The algorithm is implemented in C code with OpenMPI. The numerical experiments demonstrated  that DID has faster convergence than the other algorithms. As the update only requires basic matrix operations, DID achieves linear scalability, which is observed in the experimental results. In the future work, DID will be applied to the cases where updating matrix $B$ is also carried out in parallel. Using the techniques introduced by \cite{hsieh2011fast} and \cite{gillis2012accelerated}, DID has the possibility to be accelerated. How to better treat sparse datasets is also a potential research direction.

\clearpage
\bibliographystyle{aaai}
\bibliography{refs}
\end{document}